\documentclass[fleqn,10pt]{wlscirep}
\usepackage[utf8]{inputenc}
\usepackage[T1]{fontenc}

\usepackage[title]{appendix}%
\usepackage{dirtytalk}
\usepackage{tabularx}

\title{Visual cognition in multimodal large language models}
\author[1,3,+,*]{Luca M. Schulze Buschoff}
\author[1,2,3,+]{Elif Akata}
\author[2]{Matthias Bethge}
\author[1,3]{Eric Schulz}
\affil[1]{Max Planck Institute for Biological Cybernetics, 72076 T\"ubingen, Germany}
\affil[2]{University of T\"ubingen, 72076 T\"ubingen, Germany}
\affil[3]{Institute for Human-Centered AI, Helmholtz Munich, 85764 Oberschleißheim, Germany}
\affil[*]{lucaschulzebuschoff@gmail.com}
\affil[+]{these authors contributed equally to this work}

\begin{abstract}
A chief goal of artificial intelligence is to build machines that think like people. Yet it has been argued that deep neural network architectures fail to accomplish this. Researchers have asserted these models' limitations in the domains of causal reasoning, intuitive physics, and intuitive psychology. Yet recent advancements, namely the rise of large language models, particularly those designed for visual processing, have rekindled interest in the potential to emulate human-like cognitive abilities. This paper evaluates the current state of vision-based large language models in the domains of intuitive physics, causal reasoning, and intuitive psychology. Through a series of controlled experiments, we investigate the extent to which these modern models grasp complex physical interactions, causal relationships, and intuitive understanding of others' preferences. Our findings reveal that, while some of these models demonstrate a notable proficiency in processing and interpreting visual data, they still fall short of human capabilities in these areas. 
Our results emphasize the need for integrating more robust mechanisms for understanding causality, physical dynamics, and social cognition into modern-day, vision-based language models, and point out the importance of cognitively-inspired benchmarks.
\end{abstract}
\begin{document}

\flushbottom
\maketitle

\thispagestyle{empty}
\section*{Introduction}
\begin{figure}[ht!]
    \centering
    \includegraphics[width=1.0\textwidth]{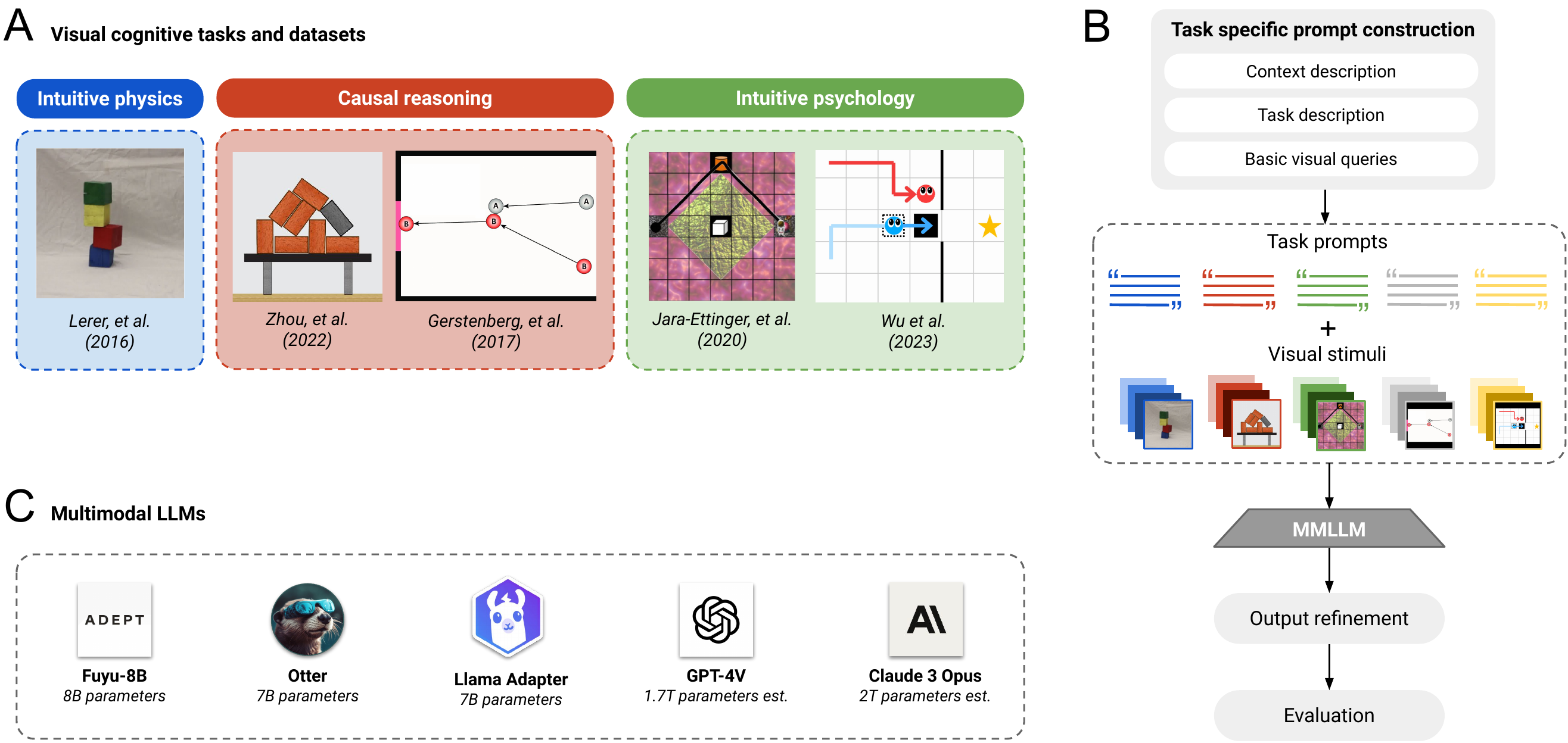}
    \caption{Overview of domains tasks, approach, and models. \textbf{A:} Example images for the different experiments. Each experiment was taken from one of three cognitive domains: intuitive physics, causal reasoning, and intuitive psychology. \textbf{B:} General approach. For every query, an image was submitted to the model, and different questions were asked about the image, i.e. we performed visual question answering. \textbf{C:} Used multi-modal large language models and their size. }
    \label{fig:overview}
\end{figure}
\newpage

People are quick to anthropomorphize, attributing human characteristics to non-human agents \cite{mitchell2019artificial}. 
The tendency to anthropomorphize has only intensified with the advent of Large Language Models (LLMs) \cite{devlin2018bert}. 
LLMs apply deep learning techniques to generate text \cite{vaswani2017attention}, learning from vast datasets to produce responses that can be startlingly human-like \cite{brown2020language}. Astonishingly, these models cannot only generate text. When scaled up to bigger training data and architectures, other, so-called ``emergent abilities'' appear \cite{bubeck2023sparks,wei2022emergent}. The current models can, for example, pass the bar exam \cite{katz2023gpt}, write poems \cite{sawicki2023power}, compose music \cite{borsos2023audiolm}, and assist in programming and data analysis tasks \cite{poldrack2023ai}. As a result, the line between human and machine capabilities is increasingly blurred \cite{kasneci2023chatgpt,bommasani2021opportunities}. People not only interact with these systems as if they were humans \cite{elkins2020can}, but they also start to rely on them for complex decision-making \cite{dell2023navigating}, artistic creation \cite{bavsic2023better}, and personal interactions \cite{akata2023playing}. It is, therefore, natural to ask: Have we built machines that think like people?

Judging whether or not artificial agents can mimic human thought is at the core of cognitive science \cite{simon1980cognitive,rumelhart1988parallel}. Therein, researchers have long debated the capabilities of artificially intelligent agents \cite{wichmann2023deep,bowers2023importance,marcus2018deep}. In a seminal paper, Lake et al. \cite{lake2017building} proposed core domains to consider when making such judgments. Published during the height of the deep learning revolution \cite{sejnowski2018deep}, the authors focused on domains that were easy for people but difficult for deep learning models: intuitive physics, causal reasoning, and intuitive psychology. 

Research on intuitive physics has studied how people perceive and interpret physical phenomena \cite{smith2018different,bates2019modeling,battaglia2012computational}. Past work on this topic has emphasized that humans possess an innate ability to predict and understand the physical properties of objects and their interactions \cite{bramley2018intuitive}, even from a young age \cite{ullman2020bayesian}, a notion sometimes summarized as a ``physics engine'' in people's heads \cite{ullman2017mind}. This understanding includes concepts such as gravity \cite{hamrick2011internal}, inertia \cite{mildenhall2001instability}, and momentum \cite{todd1982visual}. Some of the most canonical tasks in this domain involve testing people's judgments about the stability of block towers \cite{battaglia2013simulation,hamrick2016inferring}. These tasks have made their way into machine learning benchmarks \cite{bakhtin2019phyre,riochet2018intphys}, where they are used to test the intuitive physical understanding of neural networks (see \cite{buschoff2023acquisition} for an overview of previous work on building models with human-like physical knowledge).

Research on causal reasoning has studied how individuals infer and think about cause-effect relationships \cite{waldmann2017oxford,cheng1997covariation,holyoak2011causal}. Past work on this topic has proposed that humans possess an intuitive capacity to infer, understand, and predict causal relationships in their environment \cite{pearl2009causality,griffiths2009theory,lagnado2007beyond,carey1995origin}, oftentimes described using Bayesian models of causal learning \cite{gopnik2004theory,lucas2010learning}. This cognitive ability encompasses recognizing patterns \cite{bramley2018time,griffiths2005structure}, inferring causes from interventions \cite{schulz2007learning,bramley2017formalizing}, and predicting future events based on hypothetical events \cite{gerstenberg2022would}. Canonical tasks in this domain often involve assessing individuals' ability to infer causal relationships, for example, when judging the responsibility of one object causing other objects' movement \cite{gerstenberg2017eye,gerstenberg2021counterfactual}. Causal reasoning remains a challenge, even for current machine learning approaches \cite{jin2023cladder,dasgupta2019causal}.

Research on intuitive psychology has explored how individuals infer, understand, and interpret social phenomena and mental states of other agents \cite{baker2014modeling,jern2015decision}. Past work on this topic has emphasized the concept that humans possess an inherent ability to infer and reason about the mental states \cite{velez2021learning,spelke2013core}, intentions, and emotions of others, often referred to as a ``theory of mind'' \cite{baker2011bayesian,frith2005theory}. This ability has been modeled as a Bayesian inference problem \cite{saxe2017formalizing,goodman2006intuitive,shum2019theory}. Canonical tasks in this domain often involve assessing individuals' capacity to predict actions based on understanding others' perspectives or intentions, such as determining agents' utility functions based on their actions in a given environment \cite{baker2017rational,zhi2022solving}. It is the subject of ongoing debates if modern algorithms show any form of intuitive psychology \cite{rabinowitz2018machine,kosinski2023theory,ullman2023large}. 

Lake et al. argued that some of these abilities act as ``start-up software,'' because they constitute cognitive capabilities present early in development.  Moreover, they proposed that these so-called ``intuitive theories'' \cite{schulz2012origins,ullman2015nature} need to be expressed explicitly using the calculus of Bayesian inference \cite{tenenbaum2011grow}, as opposed to being learned from scratch, for example, via gradient descent. However, with the increase in abilities of current neural networks, in particular LLMs, we pondered: Can LLMs, in particular vision LLMs, sufficiently solve problems from these core domains?

To address this question, we took canonical tasks from the domains of intuitive physics, causal reasoning, and intuitive psychology that could be studied by providing images and language-based questions. We submitted them to some of the currently most advanced LLMs. In order to evaluate whether the LLMs show human-like performance in these domains, we follow the approach outlined by Binz and Schulz \cite{binz2023using}: we treat the models as participants in psychological experiments. This allows us to draw direct comparisons with human behavior on these tasks. Since the tasks are designed to test abilities in specific cognitive domains, this comparison allows us to investigate in which domains multimodal large language models perform similar to humans, and in which they don't. Our results showed that these models can, at least partially, solve these tasks. In particular, two of the largest currently available models, OpenAI's Generative Pre-trained Transformer (GPT-4) and Anthropic's Claude-3 Opus managed to perform robustly above chance in two of the three domains. Yet crucial differences emerged. First, none of the models matched human-level performance in any of the domains. Secondly, none of the models fully captured human behavior, leaving room for domain-specific models of cognition such as the Bayesian models originally proposed for the tasks.

\newpage
\subsection*{Related work}
We are not the first to assess LLMs' reasoning abilities \cite{huang2022towards,sawada2023arb,wei2022chain}. Previous studies have focused, among others, on testing LLMs' cognitive abilities in model-based planning \cite{binz2023using}, analogical reasoning tests \cite{webb2023emergent}, exploration tasks \cite{coda2023inducing}, systematic reasoning tests \cite{eisape2023systematic,hagendorff2023human}, psycholinguistic completion studies \cite{ettinger2020bert}, and affordance understanding problems \cite{jones2022distrubutional}. In this sense, our contribution can be seen as a part of a larger movement in which researchers use methods from the behavioral sciences to understand black box machine learning models \cite{rahwan2019machine,schulz2020computational,rich2019lessons}. However, most of the previous studies did not investigate multi-modal LLMs but rather remained in the pure language domain. Although there are recent attempts to investigate vision LLMs cognitive features, including their reaction to visual illusions \cite{zhang2023grounding} as well as how they solve simple intelligence tasks \cite{mitchell2023comparing}, we are the first to investigate the proposed core components of cognition in these models.  

Previous work has also looked at how LLMs solve cognitive tasks taken from the same domains that we have looked at. In intuitive physics, Ze{\v{c}}evi{\'c} et al. \cite{zevcevic2023causal} found that LLMs performed poorly in a task using language descriptions of physical scenarios. Zhang and colleagues \cite{zhang2023grounded} extracted programs from text produced by large language models to improve their physical reasoning abilities. Finally, Jassim and colleagues \cite{jassim2023grasp} proposed a novel benchmark for evaluating multimodal LLMs' understanding of situated physics.  In causal reasoning, Binz and Schulz \cite{binz2022using} showed that GPT-3 failed at simple causal reasoning experiments, while Kosoy et al. \cite{kosoy2022towards} showed that LLMs cannot learn human-like causal over-hypotheses. In research on intuitive psychology, Kosinsky argued that theory of mind might have emerged in LLMs \cite{kosinski2023theory} which has been criticized other researchers \cite{ullman2023large}. Akata et al. showed that GTP-4 plays repeated games very selfishly and could not pick up on simple conventions such as alternating between options \cite{akata2023playing}.  Finally, Gandhi and colleagues \cite{gandhi2023understanding} proposed a framework for procedurally generating Theory of Mind evaluations and found that GPT4's abilities mirror human inference patterns, though less reliable, while all other LLMs struggled. 

Many of the past studies on LLMs have fallen risk of appearing in new models' training set. Recent work has recognized this issue and, in turn, evaluated language models on many problem variations to minimize training set effects \cite{srivastava2022beyond}. Our work differs from these approaches as current models could not have just memorized solutions to the given problems because these problems require higher level reasoning. Furthermore, the human data and ground truth are most commonly stored in additional data files, which first have to be extracted and matched to the respective images in order to be used for model training. Since this requires data wrangling that can't easily be automated and the number of stimuli to gain is so small, it is extremely unlikely that these stimuli together with the ground truth were entered into the training set of any of the investigated models.
    
\section*{Results}\label{sec2}

We tested five different models on three core components for human-like intelligence as outlined by Lake et al. (\cite{lake2017building}; see Fig.~\ref{fig:overview}A). The models we used are vision large language models, which are multimodal models that integrate image processing capabilities into large language models (\cite{baltruvsaitis2018multimodal,reed2016generative}; see Fig.~\ref{fig:overview}C). These models allow users to perform visual question answering \cite{wu2017visual,manmadhan2020visual}: users can upload an image and ask questions about it, which the model interprets and responds to accordingly.   

To test the three core components, we used tasks from the cognitive science literature that could be studied in vision LLMs via visual question answering. For every task, we queried the visual reasoning abilities of the LLMs with tasks of increasing complexity. First, we asked about simple features of the shown images such as the background color or the number of objects shown. Afterward, we submitted questions taken from the cognitive science experiments. We report results based on comparisons to the ground truth as well as the different models' matches to human data.

\newpage

    \begin{figure*}[t!]
        \centering
        \includegraphics[width=1.0\textwidth]{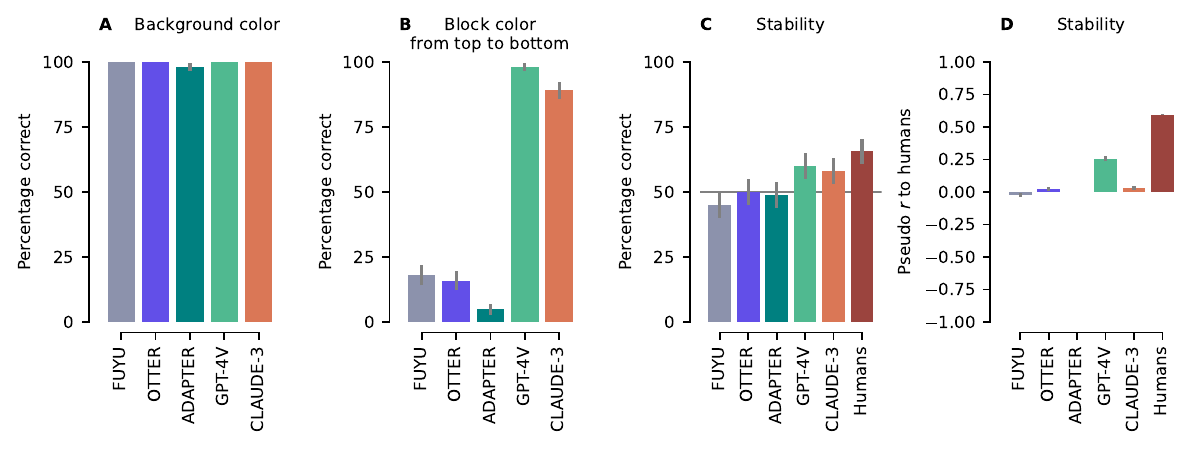}
        \caption{Results for five vision large language models for tasks of increasing complexity given images of real block towers from Lerer et al. \cite{lerer2016learning}. We first ask for the background color in the image (A), then the color of blocks from top to bottom (B), and finally a binary stability rating for the block towers (C). The last plot shows the square root of the $R^2$ value for the Bayesian logistic mixed effects regression between models and human subjects (D). Error bars in plots A - C are given by the standard deviation of a binomial distribution. Error bars in plot D are given by the square root of the 95\% percentiles for the Bayesian $R^2$ value.}
        \label{fig:phys}
    \end{figure*}

\subsection*{Intuitive physics: Block towers}\label{subsec2.1}
\vspace{0.5cm}
To test the intuitive physics capabilities of the different LLMs, we used photographs depicting wooden block towers from Lerer et al. \cite{lerer2016learning} (see Fig.~\ref{fig:phys_example} in the Appendix for an example). We first aked models to determine the background color of the image. All four models achieved almost perfect accuracy (see Fig.~\ref{fig:phys}A). We then asked models to state the color of blocks from top to bottom. Here, the performance of most models except for GPT-4V and Claude-3 deteriorated (see Fig.~\ref{fig:phys}B). Please note that the first two tasks are fairly trivial for humans and we would expect human performance to be at 100\% (the background color is always white and images featured either 2, 3, or 4 blocks in primary colors). 

To test the models' physical reasoning abilities, we asked them to give a binary stability judgment of the depicted block towers. Here, only GPT-4V and Claude-3 performing slightly above chance (see Fig.~\ref{fig:phys}C, for GPT-4V Fisher's exact test yielded an odds ratio of $2.597$ with a one-sided p-value of $0.028$). None of the other models performed significantly above chance (the second best performing model, Claude-3, had an odds ratio of $2.016$, with a one-sided p-value of  of $0.078$). Human subjects were also not perfect but showed an average accuracy of $65.608$\%.

Finally, we determined the relationship between models' and humans' stability judgments using a Bayesian logistic mixed effects regression. We compute a Bayesian $R^2$ for each regression model based on draws from the modeled residual variances \cite{gelman2019r}. We then take the square root of this Bayesian $R^2$ and multiply it with the sign of the main regression coefficient to arrive at a pseudo $r$ value. Around this pseudo $r$ value we plot the square root of the 95\% percentiles for the $R^2$ value (see Fig.~\ref{fig:phys}D). We found that GPT-4V was the only model that showed a relation to human judgments, with a regression coefficient of $1.15$ ($95\% CI = [1.04, 1.27]$) and an $R^2$ value of $0.066$. However, the regression coefficient between individual humans and the mean over humans was still larger with a coefficient of $1.46$ ($95\% CI = [1.41, 1.52]$) and an $R^2$ value of $0.354$.

\newpage

    \begin{figure*}[t!]
        \centering  
        \includegraphics[width=1.0\textwidth]{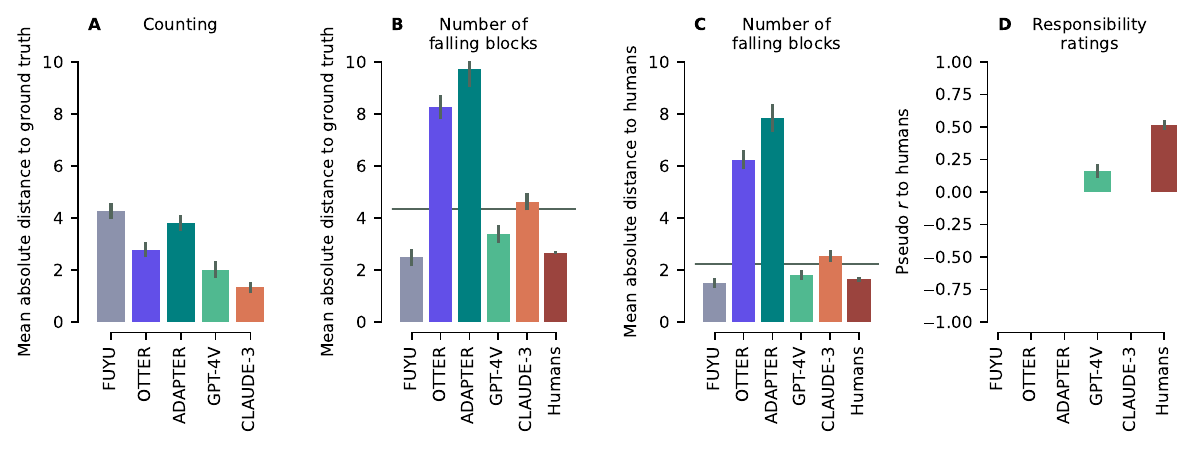}
        \caption{Results for causal reasoning experiment from Zhou et al. \cite{zhou2023mental}. We first ask for the number of blocks in the image (A), then the number of blocks that would fall if a specific block is removed (B and C), and finally a rating between 0 and 100 for how responsible a specific block is for the stability of the tower (D). For the responsibility ratings, all LLMs except for GPT-4V give constant ratings: Fuyu and Claude-3 always respond with 100, while Otter and LLaMA-Adapter V2 always respond with 50. Error bars in plots A - C are given by the standard error of the mean, while the error bar plot D is given by the square root of the 95\% percentiles for the Bayesian $R^2$ value.}
        \label{fig:caus}
    \end{figure*}

\subsection*{Causal reasoning: Jenga}\label{subsec2.2}
\vspace{0.5cm}
To test the models' causal reasoning capabilities, we used synthetic images from Zhou et al. \cite{gerstenberg2017faulty, zhou2023mental}, which depicted block towers that were stable but might collapse if one of the blocks was removed (see Fig.~\ref{fig:caus_example} in the Appendix for an example). We started by asking the models to count the blocks in the image. The images in this task displayed a larger number of blocks (ranging from 6 to 19), which made the basic counting task significantly more challenging than in the previous section. Models' responses approximated the ground truth, albeit rarely matching it exactly. Therefore, we report the mean absolute distance to the ground truth instead of the percentage of correct answers (see Fig. \ref{fig:caus}A). The models' performance highlighted the challenging nature of this task, with the best performing model (Claude-3) still being on average more than one block off. 

We continued by querying the models for the number of blocks that would fall if a specific block was removed from the scene (see Figures \ref{fig:caus}B and \ref{fig:caus}C). We established a baseline performance represented by a horizontal line in Fig.~\ref{fig:caus}B-C, which corresponds to a random agent that gave the mean between 0 and the number of blocks in each image as its' prediction, essentially behaving like a uniform distribution over the possible number of blocks that could fall. Notably, both GPT-4V and Fuyu-8B surpassed the random baseline, their performance levels being close to the human results reported in Zhou et al. \cite{zhou2023mental}, which is depicted by the rightmost bar in the plot. However, GPT-4V still diverges significantly from the average over human subjects ($t(42)=2.59$, $p<.05$).

Finally, we asked the models to rate the responsibility of a specific block for the stability of the other blocks (see Fig.~\ref{fig:caus}D). Notably, all models except for GPT-4V gave constant ratings for this task (Fuyu and Claude-3 always responded with 100, while Otter and LLaMA-Adapter V2 always responded with 50). The regression coefficient for GPT-4V with human values is $0.16$ (95\% CI = [0.10, 0.21]) with an $R^2$ value of $0.027$. The human-to-human regression has a coefficient of $0.54$ (95\% CI = [0.45, 0.63]) and an $R^2$ value of $0.268$. 

\newpage 

    \begin{figure*}[t!]
        \centering  
        \includegraphics[width=1.0\textwidth]{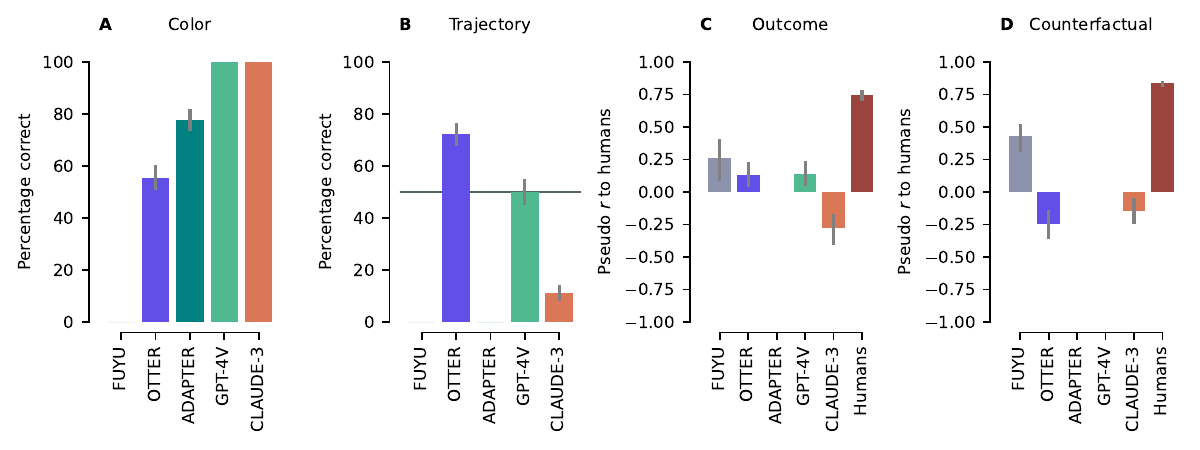}
        \caption{Results for causal reasoning experiment taken from Gerstenberg et al. \cite{gerstenberg2017eye}. We first ask for the background color in the image (A), then the direction of ball movement (B), a judgement between 0 and 100 on whether ball \protect\say{B} goes through the gate (C), and finally a counterfactual judgement between 0 and 100 on whether ball \protect\say{B} would have gone through the gate, had ball \protect\say{A} not been present in the scene (D). The error bars in plots C and D are given by the the square root of the 95\% percentiles for the Bayesian $R^2$.}
        \label{fig:caus2}
    \end{figure*}

\subsection*{Causal reasoning: Michotte}\label{subsec2.2}
\vspace{0.5cm}
For the second test for causal reasoning abilities, we ran an experiment from Gerstenberg et al. \cite{gerstenberg2017eye} that is based on the classic Michotte launching paradigm \cite{michotte1963causality}. It uses simple synthetic 2D depictions of two balls labelled \say{A} and \say{B} with arrows showing their trajectories in front of a white background. We started by asking the models to determine the background color of the image (see Fig. \ref{fig:caus2}A). Most models perform fine with the exception of Fuyu, which always answers \say{pink} (likely since pink is mentioned as the color of the gate in the prompt). Then, we asked models to infer the trajectory of ball movement. This proved challenging for most models (see Fig. \ref{fig:caus2}B), which is surprising given that the prompt explicitly mentions that the arrows in the stimuli depict the trajectory of the balls and the balls always move from right to left. 

We then queried the models for their agreement on a scale from 0 to 100 to the following questions: either \say{Ball B went through the middle of the gate} (if ball A entered the gate) or \say{Ball B completely missed the gate} (if ball B missed the gate) (see Fig. \ref{fig:caus2}C). No model performs close to the human results reported in Gerstenberg et al. \cite{gerstenberg2017eye}. The best performing model is Fuyu with a regression coefficient of $0.26$ (95\% CI = [-0.08, 0.61]) and an $R^2$ value of $0.067$. Interestingly, Claude-3 shows a negative relationship with human judgements, with a regression coefficient of $-0.22$ (95\% CI = [-0.39, -0.06]) and with an $R^2$ value of $0.076$. The human-to-human regression coefficient is $0.85$ (95\% CI = [0.69, 1.03]) with an $R^2$ value of $0.556$. 

Finally, we asked the models for their agreement on a scale from 0 to 100 to the counterfactual question if \say{Ball B would have gone through the gate had Ball A not been present in the scene} (see Fig. \ref{fig:caus2}D). Notably, the closed source models perform worse than some open source models for both tasks. Here, Fuyu is again the best performing model with a regression coefficient of $0.42$ ($95\% CI = [0.28, 0.57])$ and an $R^2$ value of $0.185$. Pseudo $r$ values for LLaMA-Adapter V2 and GPT-4V are missing, since the former only gave non-valid answers and latter always responded with 100. The human-to-human regression coefficient is $0.85$ (95\% CI = [0.76, 0.93]) with an $R^2$ value of $0.698$.
\newpage 

    \begin{figure*}[t!]
        \centering
        \includegraphics[width=1.0\textwidth]{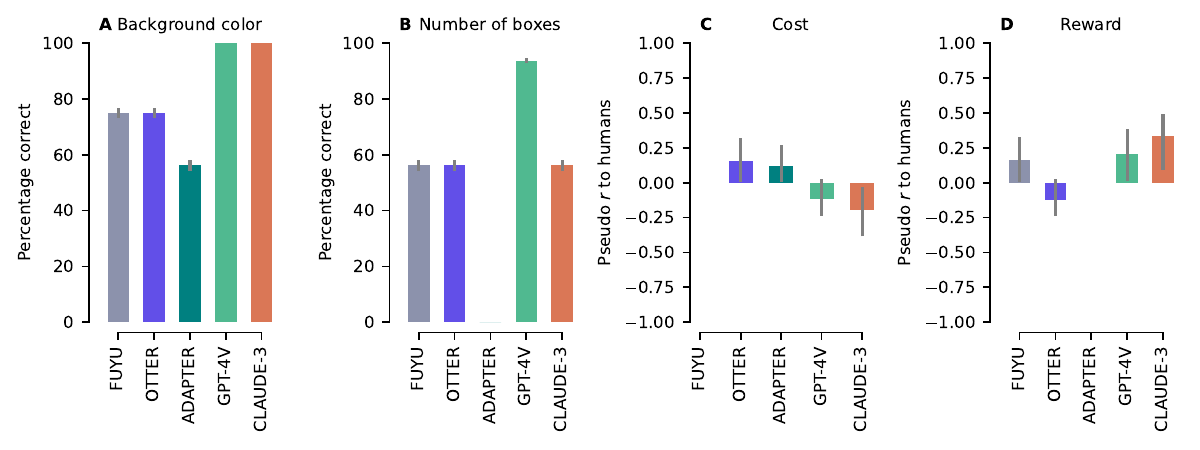}
        \caption{Results on tasks for intuitive psychology taken from Jara-Ettinger et al. \cite{jara2020naive}. Again, we first ask for the background color (A) and the number of boxes in the scene (B). Models are then asked to make inferences about the costs and rewards in an environment depending on the path an agent has taken (C and D). Regression coefficients for Fuyu and LLaMA-Adapter V2 are missing as they always responded with constant ratings for either cost or reward questions. Error bars in plot A are given by the standard deviation of a binomial distribution, while the error bars in plots C and D are given by the square root of the 95\% percentiles for the Bayesian $R^2$ value.}
        \label{fig:psych}
    \end{figure*}

\subsection*{Intuitive psychology: Astronaut}\label{subsec2.3}
\vspace{0.5cm}
As a first test for the intuitive psychology understanding of the different LLMs we used synthetic images depicting an astronaut on a colored background from Jara-Ettinger et al. \cite{jara2020naive} (see Fig.~\ref{fig:phys_example} in the Appendix for an example). The images featured different terrains and care packages. Depending on which terrain the astronaut crossed or which care package they chose to pick up, it was possible to infer the costs associated with the terrains and rewards associated with the care packages. 

Again, we first tasked models with determining the background color of the images. Here, the performance of the models was worse compared to the intuitive physics data set (see Fig. \ref{fig:psych}A), which might be due to the fact that the background color here was not uniform (see Fig. \ref{fig:psych_example}). We then asked models to count the number of care packages in the scene. Most models except for GPT-4V struggled here (see Fig. \ref{fig:psych}B). 

Afterwards we asked them to infer the costs associated with the different terrains (see Fig. \ref{fig:psych}C) and the rewards associated with different care packages (see Fig. \ref{fig:psych}D). All models only showed weak relations with the average over human subjects in their judgments about the costs and rewards associated with the environment. The regression coefficients of the models with the z-scaled mean over human subjects ranged from $-0.24$ to $0.16$ with $R^2$ values between $0.025$ to $0.04$ for cost questions, and from $-0.02$ to $0.39$ (Claude-3, 95\% CI = [0.11, 0.66]) with $R^2$ values between $0.015$ and $0.110$ for reward questions.

\newpage

    \begin{figure*}[t!]
        \centering
        \includegraphics[width=1.0\textwidth]{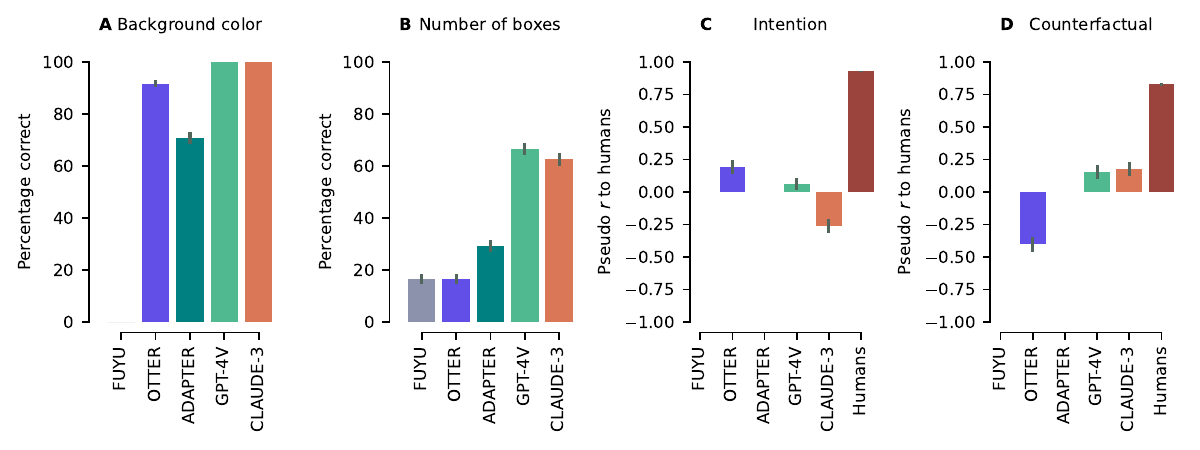}
        \caption{Results on the new intuitive psychology data set from Wu et al. \cite{wu2023responsibility}. We first ask for the background color in the image (A), then the number of boxes in the scene (B), a judgement between 0 and 100 on whether an agent in the scene tried to hinder the other agent (C), and finally a counterfactual judgement between 0 and 100 on whether an agent in the scene would have succesfully reached the goal, had the other agent not been present (D). The error bars in plots C and D are given by the square root of the 95\% percentiles for the Bayesian $R^2$ value.}
        \label{fig:agents_main}
    \end{figure*}
    
\subsection*{Intuitive psychology: Help or Hinder}\label{subsec2.3}
\vspace{0.5cm}

The new intuitive psychology dataset we added is taken from Wu et al. \cite{wu2023responsibility}. This task shows a simple 2D depiction of two agents in a grid environment. On each time step, the agents can move up, down, left, right, or stay in place, but cannot move through walls or boxes. The red agent has the objective of reaching a star in 10 time steps. If the agent runs out of time they fail. The blue agent has the objective of either helping or hindering the red agent by pushing or pulling boxes around. 

We first asked models to determine the background color in the scene and to determine the number of boxes in the scene (see panels A and B in Fig. \ref{fig:agents_main}). The closed source models are able to perfectly determine the background color (always white) but they nonetheless struggle with determining the number of boxes in the scene (always either 1, 2, or 3). We then asked the models whether the blue agent tried to help or hinder the red agent (see Fig. \ref{fig:agents_main}C). Here, Otter shows the highest regression coefficient with human answers with $0.19$ (95\% CI = [0.13, 0.25]) with an $R^2$ value of $0.038$. Claude-3 shows a negative relationship with human answers with a coefficient of $-0.25$ (95\% CI = [-0.31, -0.20]) and an $R^2$ value of $0.066$. No model showed coefficients even close to the human-to-human coefficient of $0.93$ (95\% CI = [0.90, 0.96]) with an $R^2$ value of $0.858$. 

Finally, we asked the model if the red agent would have succeeded in reaching the star, had the blue agent not been there. We show the square root of the $R^2$ for the Bayesian linear mixed effects regression with 95\% percentiles in Fig. \ref{fig:agents_main}D. Interestingly, the results here flip, with Otter now showing a stronger negative relationship with a coefficient of $-0.40$ (95\% CI = [-0.47, -0.33]) and an $R^2$ value of $0.161$ (this makes sense, since this task is essentially a counterfactual simulation question similar to Fig.~\ref{fig:caus2}D, where Otter already showed a negative relation to human judgements). GPT-4V and Claude-3 both show small positive regression coefficients with humans answers: $0.15$ (95\% CI = [0.09, 0.21]) with an $R^2$ value of $0.025$, and $0.17$ (95\% CI = [0.11, 0.23]) with an $R^2$ value of $0.032$, respectively. Again, no model coefficient is close to the human-to-human coefficient of $0.83$ (95\% CI = [0.80, 0.87]) with an $R^2$ value of $0.688$.

\newpage
\section*{Discussion}
\vspace{0.5cm}
We started by asking whether, with the rise of modern large language models, researchers have created machines that -- at least to some degree -- think like people. To address this question, we took four recent multi-modal large language models and probed their abilities in three core cognitive domains: intuitive physics, causal reasoning, and intuitive psychology. 

In intuitive physics and causal reasoning, the models managed to solve some of the given tasks and GPT-4V showed a slight match with human data. However, while they performed well in some tasks, the models did not show a conclusive match with human data for the causal reasoning experiments. Finally, in the intuitive psychology tasks, none of the models showed a strong match with human data. Thus, an appropriate answer to the question motivating our work would be ``No.'', or -- perhaps more optimistically -- ``Not quite.''

Although we have tried our best to give all models a fair chance and set up the experiments in a clean and replicable fashion, some shortcomings remain that should be addressed in future work. First of all, we have only tested a handful of multi-modal models on just three cognitive domains. While we believe that the used models and tasks provide good insights into the state-of-the-science of LLMs' cognitive abilities, future studies should look at more domains and different models to further tease apart when and why LLMs can mimic human reasoning. For example, it would be interesting to see if scale is the only important feature influencing model performance \cite{sutton2019bitter,kaplan2020scaling}. Currently, our evidence suggests that even smaller models, for example Fuyu with its 8 billion parameters, can sometimes perform as well as GPT-4V in some tasks. Additionally, we applied all models out of the box and without further fine-tuning. Future studies could attempt to fine-tune multi-modal LLMs to better align with cognitive data \cite{binz2023turning} and assess if this improves their reasoning abilities more generally. 

Another shortcoming of the current work is the simplicity of the used stimuli. While the block towers used in our first study were deliberately designed to be more realistic \cite{lerer2016learning} than commonly used psychological stimuli \cite{battaglia2013simulation}, this was not true for the experiments in the other two domains. For the intuitive psychology experiments, in particular, we would expect the models to perform better if the stimuli contained more realistic images of people, which has been shown to work better in previous studies \cite{ju2022prompting}. Interestingly, using more realistic stimuli can also change people's causal judgments \cite{meding2020phenomenal}; how realistic stimuli used in cognitive experiments should be, remains an open question \cite{allen2023using}. 

On a related point, we only used static images in our current experiments, which severely limits the breadth and level of detail of the questions we could ask. For example, some of the most canonical tasks investigating people's causal reasoning abilities involve videos of colliding billiard balls \cite{gerstenberg2017eye}. As future large language models will likely be able to answer questions about videos \cite{maaz2023video}, these tasks represent the next frontier of cognitively-inspired benchmarks. 

For the comparisons to human data, we used the participant data collected in the original studies for all experiments except for the intuitive physics task and assessed the correspondence between models and this data via a Bayesian mixed effects regression and $R^2$ values. Future work could expand on this approach by collecting new data from human subjects choosing which of the model's judgments they prefer. This could lead to a more detailed comparison, similar to what has been proposed to discriminate among deep learning models for human vision \cite{golan2020controversial} and language \cite{golan2023testing}.

A crucial weakness of most studies using large language models is that they can be sensitive to specific prompts \cite{reynolds2021prompt,strobelt2022interactive,webson2021prompt}. While we have attempted to use prompts that elicited good behavior, thereby giving LLMs a chance to perform well, future work could try to further optimize these prompts using available methods \cite{liu2023pre,gu2023systematic,coda2023meta}, while also assessing how the models respond to paraphrased versions of the same tasks. We show a first exploratory analysis for the effects of response constraints and context complexity on correlations to human behavior in the intuitive psychology astronaut task in the Appendix. While response constraints and context complexity both influence model outputs, we also find that small variations to prompts on a character level can impact model behavior, likely due to tokenization. Taken together, this shows that evaluations of visual large language models are not only dependent on the specific models and experiments used, but also the prompts and likely even how these prompts are tokenized. While it could be possible to further engineer the used prompts, we believe that our current approach was sufficient to showcase these models' abilities.


Our work has shown that multimodal LLMs have come a long way, showing some correspondence to human behavior and often performing above chance. Moreover, machine learning researchers have put forward various ideas about how to close the remaining gap between humans and machines \cite{geirhos2021partial}, including self-supervised learning \cite{balestriero2023cookbook}, translating from natural into probabilistic languages \cite{wong2023word}, or grounding LLMs in realistic environments \cite{carta2023grounding}. This continuous evolution in models' capabilities necessitates a reevaluation of the metaphors and tools we use to understand them. We believe that cognitive science can offer tools, theories, and benchmarks to evaluate how close we have come to \say{building machines that learn and think like people}.

\newpage
\section*{Methods}
\vspace{0.5cm}
    \subsection*{Code}
    The open-source models were installed per the instructions on their related Github or Huggingface repositories and evaluated on a Slurm-based cluster with a single A100. For the results reported as GPT-4V, we used the public ChatGPT interface and the OpenAI API, specifically the November 2023 release of \texttt{gpt4-vision-preview} model which is available via the completions endpoint. For CLAUDE-3 we used the Anthropic API. Code for replicating our results is available on GitHub (\href{https://github.com/lsbuschoff/multimodal}{github.com/lsbuschoff/multimodal}). All models were evaluated in Python using PyTorch \cite{paszke2019pytorch}. Additional analyses were carried out using NumPy \cite{harris2020array}, Pandas \cite{reback2020pandas}, and SciPy \cite{2020SciPy-NMeth}. Matplotlib \cite{Hunter:2007} and Seaborn \cite{Waskom2021} were used for plotting. Bayesian mixed effects models were computed using brms \cite{buerkner2017brms} in R \cite{rteam2021r}. 
    
    \subsection*{Models}
    
        \subsubsection*{Open-source}
        Fuyu is an 8B parameter multi-modal text and image decoder-only transformer. We used the Huggingface implementation with standard settings and without further finetuning (available \href{https://huggingface.co/adept/fuyu-8b}{here}). The maximum number of generated tokens was set to 8 and responses were parsed by hand. Otter is a multi-modal LLM that supports in-context instruction tuning and it is based on the OpenFlamingo model. We used the Huggingface implementation of \texttt{OTTER-Image-MPT7B} (available \href{https://huggingface.co/luodian/OTTER-Image-MPT7B}{here}), again with standard settings and without finetuning. The maximum number of generated tokens was left at 512 and responses were parsed by hand. For LLaMA-Adapter V2, which adds adapters into LLaMA's transformer in order to turn it into an instruction-following model, we used the GitHub implementation of \texttt{llama-adapter-v2-multimodal7b} with standard settings and again without further finetuning (available \href{https://github.com/OpenGVLab/LLaMA-Adapter/tree/main/llama_adapter_v2_multimodal7b}{here}). The maximum number of generated tokens was left at 512 and responses were parsed by hand.

        \subsubsection*{Closed-source}
        We initially queried GPT-4V through the ChatGPT interface, since the OpenAI API was not publicly available at the outset of this project. The Intuitive Psychology task responses were collected using the \texttt{gpt4-vision-preview} model variant after its November 2023 release in the API. We set the maximum number of generated tokens for a given prompt to 1 to get single numerical responses. All other parameters were set to their default values. Note that this model does not currently feature an option for manually setting the temperature, and the provided documentation does not specify what the default temperature is. We query Claude-3 using the Anthropic API. We use the model version \texttt{claude-3-opus-20240229} with a temperature of zero and the maximum number of new tokens between 3 and 6 depending on the task. 

\vspace{1cm}
\subsection*{Datasets}
    \subsubsection*{Intuitive physics: Block towers}
    \label{sec:datasetintphys}
    We tested the intuitive physical understanding of the models using images from Lerer et al. \cite{lerer2016learning}. The photos depict a block tower consisting of colored wooden blocks in front of a white fabric (see Fig. \ref{fig:phys_example} for an example). The images are of size 224 x 244. In the data set, there are a total of 516 images of block towers. We tested the models on 100 randomly drawn images. We first tested the models on their high-level visual understanding of the scenes: we tasked them with determining the background color and the number of blocks in the image. In order to test their physical understanding, we tested them on the same task as the original study: we asked them to give a binary rating on the stability of the depicted block towers. For the first two tasks, we calculated the percentage of correct answers for each of the models. For the third task, we calculated a Bayesian linear mixed effects regression between human and model answers.

    Due to the limited sample size of the original human experiment, we reran the human experiment by Lerer et al. \cite{lerer2016learning} on Prolific with 107 subjects (55 female and 52 male native English speakers with a mean age of $27.73$ (sd = $4.21$). Subjects first saw an example trial, followed by 100 test images. In a 2AFC paradigm, subjects were asked if the block tower in a given image was stable or not stable. They were paid £1.5 and the median time they took to complete the experiment was 08:08 minutes, making the average base reward £11.07 per hour. Additionally, they received a bonus payment of up to £1 depending on their performance (1 cent for each correct answer)

    \newpage
    \subsubsection*{Causal reasoning: Jenga}
    For the first causal reasoning experiment we used images from Zhou et al. \cite{zhou2023mental}. The images show artificial block stacks of red and gray blocks on a black table (see Fig. \ref{fig:caus_example} for an example). The data set consists of 42 images on which we tested all models. We again first tested the models on their high-level visual understanding of the scene and therefore tasked them with determining the number of blocks in the scene. The ground truth number of blocks in the scenes ranged from 6 to 19. Since this task is rather challenging due to the increased number of blocks, we do not report the percentage correct as for the intuitive physics data set but the mean over the absolute distance between model predictions and the ground truth for each image (see Fig. \ref{fig:caus}A).
    
    To test the causal reasoning of the models we adopted the tasks performed in the original study \cite{zhou2023mental, gerstenberg2017faulty}. We asked models to infer how many red blocks would fall if the gray block was removed. For this condition, Zhou et al. collected data from 42 participants. We again report the absolute distance between model predictions and the ground truth for each image (see Fig. \ref{fig:caus}B). We calculate a random baseline which uses the mean between 0 and the number of blocks for each specific image as the prediction. We also ask the models for a rating between 0 and 100 for how responsible the gray block is for the stability of the tower. Here, data for 41 human participants was publicly available. For both, the number of blocks that would fall if the gray block was removed, and its' responsibility for the stability of the tower, we calculate the mean Pearson correlation to human subjects from the original study (see Fig. \ref{fig:caus}C). 

    \subsubsection*{Causal reasoning: Michotte}
    For the second test for causal reasoning abilities, we used a task from Gerstenberg et al. \cite{gerstenberg2017eye}. It features 18 images which show a 2D-view of two balls and their trajectories on a flat surface (see Fig. \ref{fig:caus2_example} for an example). This experiment is a variation of the classic Michotte launching paradigm \cite{michotte1963causality}, used to test visual causal perception. We again first tested the models on their high-level visual understanding of the scene: we first asked them to determine the background color (see Fig. \ref{fig:caus2}A) and then the direction of ball movement (see Fig. \ref{fig:caus2}B) from the two options \say{left to right} or \say{right to left} (the balls always moved from right to left).
        
    To test the causal reasoning of the models we adopted the tasks performed in the original study. We asked models about the actual outcome of the scene: "Did ball A enter the gate?". As in the original experiments, models had to indicate their agreement with this statement on a scale form 0 (not at all) to 100 (completely). We then also ask the counterfactual question: "Would ball A have entered the gate had it not collided with ball B?". The original authors collected the responses of 14 subjects in the \say{outcome} condition and 13 subjects in the \say{counterfactual} condition. We here report the regression between model and human responses (see Fig. \ref{fig:caus2}C-D).
    
    \subsubsection*{Intuitive psychology: Astronaut}
    To test the intuitive psychology of the different LLMs, we used stimuli from Jara-Ettinger et al. \cite{jara2020naive}. This part consisted of three different experiments each consisting of 16, 17, and 14 images showing a 2D depiction of an astronaut and care packages in different terrains (see Fig. \ref{fig:psych_example_all} for an example). In order to check their high-level understanding of the images, we again asked the models to determine the background color of the images. Since this background color is not uniform, we counted both ``Pink'' and ``Purple'' as correct answers. We report the percentage of correct answers for the background color in Fig. \ref{fig:psych}A.

    In accordance with the original study, analyses for the intuitive psychological capabilities of the models are split into cost questions (passing through a terrain is associated with a cost for the agent) and reward questions (collecting a care package yields some sort of reward for the agent). We pooled cost and reward questions over all three experiments and reported the mean Pearson correlation with the data of 90 human subjects collected by Jara-Ettinger et al. \cite{jara2020naive} (see Figs. \ref{fig:psych}B and \ref{fig:psych}C). This heuristic calculates the costs and rewards associated with the environment from the amount of time an agent spends in each terrain and which care package it collects. 

    \subsubsection*{Intuitive psychology: Help or Hinder}
    The second intuitive psychology experiment is taken from Wu et al. \cite{wu2023responsibility}. It consists of 24 images showing a 2D depiction of two agents in a grid world (see Fig. \ref{fig:psych2_example} for an example). In order to check the models' basic understanding of the images, we again asked the models to determine the background color of the images and the number of boxes in the scene. We report the percentage of correct answers for both tasks in Fig. \ref{fig:agents_main}A-B.
    
    We then asked the models whether the blue agent tried to help or hinder the red agent on a scale from \say{“definitely hinder RED” (0) to “definitely help RED” (100) with the midpoint “unsure” (50).}. We show the regression to human judgements in Fig. \ref{fig:agents_main}C. Finally, we asked the model the counterfactual question if the red agent would have succeeded in reaching the star had the blue agent not been there on a scale from \say{“not at all” (0) to “very much” (100)?}. The original authors collected the responses of 50 subjects for each of the two conditions (\say{intention} and \say{counterfactual}). We show the mixed linear regression coefficients between model and human answers for all models with 95\% credible intervals in Fig. \ref{fig:agents_main}D. 

\clearpage

\section*{Data availability}
All data used in our experiments is available on GitHub (\href{https://github.com/lsbuschoff/multimodal}{github.com/lsbuschoff/multimodal}). We have used subsets of openly available data sets from Lerer et al. (\href{https://github.com/facebookarchive/UETorch/issues/25#issuecomment-235688223}{https://github.com/facebookarchive/UETorch/issues/25\#issuecomment-235688223}, \cite{lerer2016learning}), Gerstenberg et al. (\href{https://github.com/tobiasgerstenberg/eye_tracking_causality}{https://github.com/tobiasgerstenberg/eye\_tracking\_causality, \cite{gerstenberg2017eye}}), Zhou et al. (\href{https://github.com/cicl-stanford/mental_jenga}{https://github.com/cicl-stanford/mental\_jenga, \cite{zhou2023mental}}), Wu et al. (\href{https://github.com/cicl-stanford/counterfactual_agents}{https://github.com/cicl-stanford/counterfactual\_agents, \cite{wu2023responsibility}}), and Jara-Ettinger et al. (\href{https://osf.io/uzs8r/}{https://osf.io/uzs8r/, \cite{jara2020naive}}).

\section*{Code availability}
All code needed to reproduce our results is available on GitHub (\href{https://github.com/lsbuschoff/multimodal}{github.com/lsbuschoff/multimodal}). We use openly available implementations of all LLMs except for GPT-4V and Claude-3. The code includes instructions on how to install and evaluate these LLMs. All prompts are listed in the supplementary information.

\bibliography{bibliography}

\section*{Acknowledgements}
We thank Mirko Thalmann and Can Demircan for helpful discussions on the statistical analysis. This work was supported by the Max Planck Society, the Volkswagen Foundation, the German Federal Ministry of Education and Research (BMBF): Tübingen AI Center, FKZ: 01IS18039A, and the Deutsche Forschungsgemeinschaft (DFG, German Research Foundation) under Germany’s Excellence Strategy – EXC 2064/1 – 390727645.

\section*{Author contributions statement}

All authors conceived the experiments. L.M.S.B. and E.A. conducted the experiments. All authors analysed the results. All authors wrote the manuscript. 

\clearpage
\begin{appendices}

\section{Example trials}

    \subsection{Intuitive physics: Block towers}
    
        \begin{figure*}[ht!]
            \centering
            \includegraphics[width=1.0\textwidth]{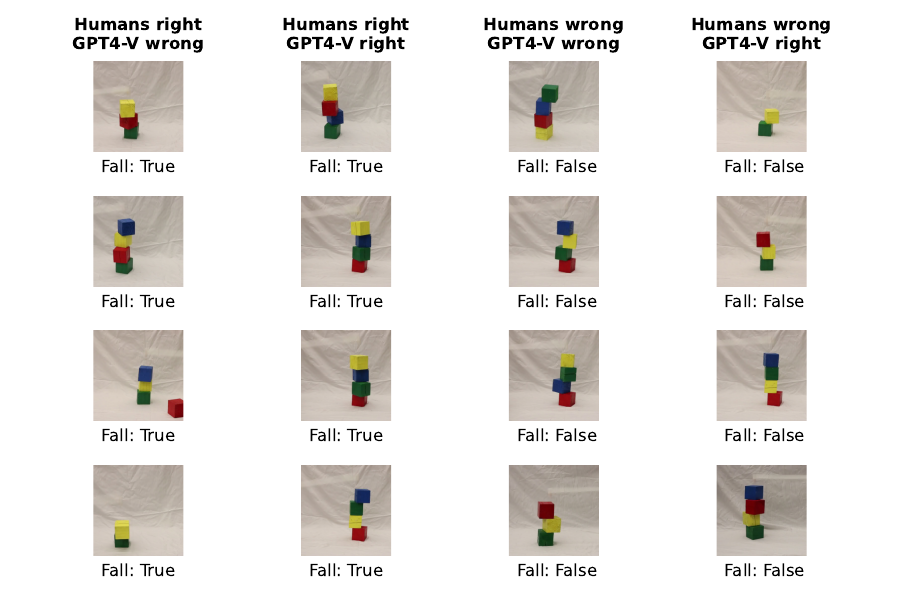}
            \caption{Example images from Lerer et al. \cite{lerer2016learning} which are either challenging or easy for Humans, GPT4-V, or both in the binary tower stability judgement task for physical intuition.}
            \label{fig:phys_example}
        \end{figure*}

    \clearpage
    \subsection{Causal reasoning: Jenga}

        \begin{figure*}[ht!]
            \centering
            \includegraphics[width=1.0\textwidth]{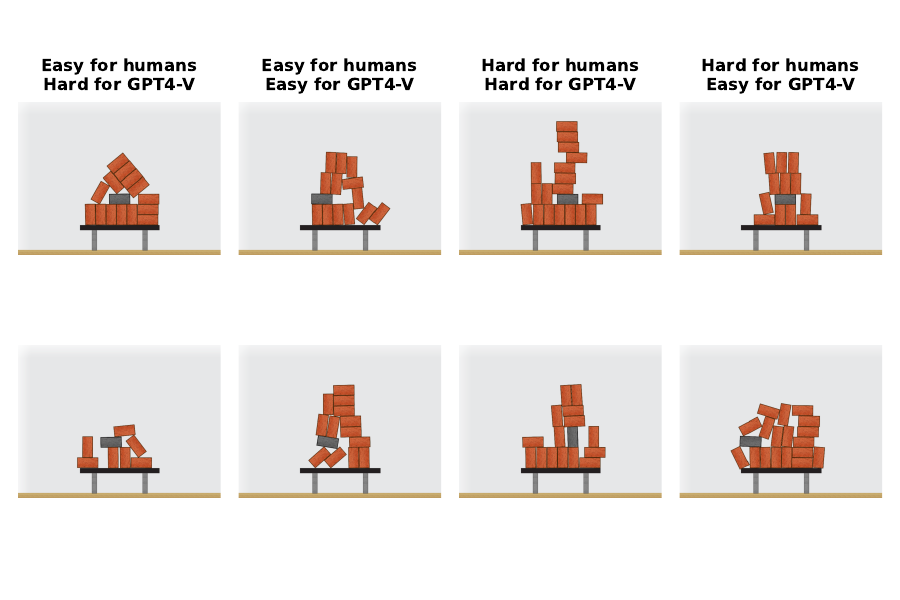}
            \vspace{-2cm}
            \caption{Example images from Zhou et al. \cite{zhou2023mental} which are either challenging or easy for Humans, GPT4-V, or both in the number of blocks that will fall task for causal reasoning. Easy here refers to images where judgements have a low average distance to the  that the ground truth. Hard refers to images where judgements have a high average distance to the ground truth.}
            \label{fig:caus_example}
        \end{figure*}

        \vspace{1cm}
        \subsection{Causal reasoning: Michotte}
    
        \begin{figure*}[ht!]
            \centering
            \includegraphics[width=1.0\textwidth]{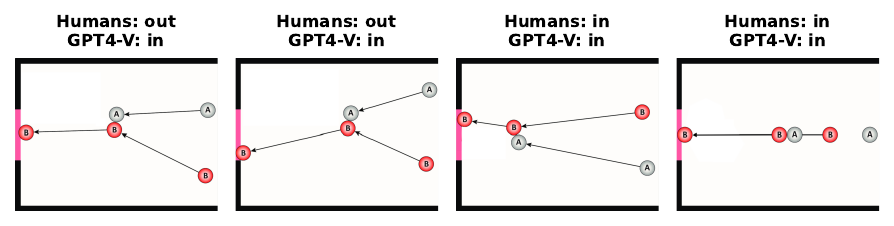}
            \caption{Example images from Gerstenberg et al. \cite{gerstenberg2017eye}. Humans and models were asked if ball “B” would have gone through the gate if ball “A” had not been present in the scene.}
            \label{fig:caus2_example}
        \end{figure*}

    \clearpage
    
    \subsection{Intuitive psychology: Astronaut}

        \begin{figure*}[ht!]
            \centering
            \includegraphics[width=1.0\textwidth]{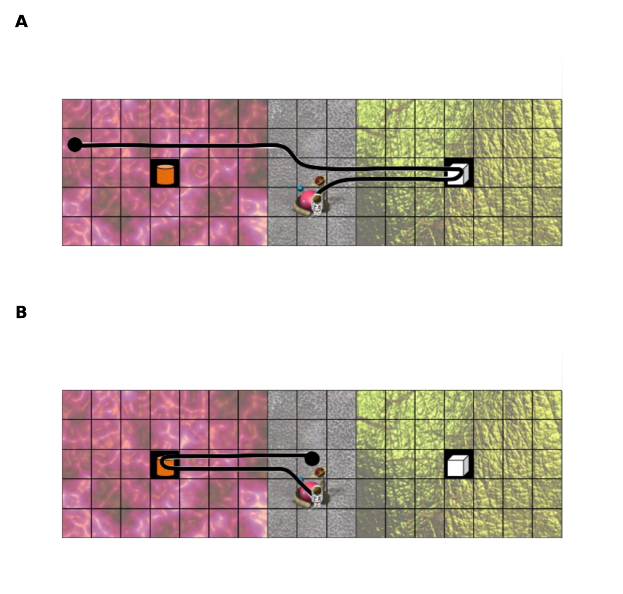}
            \caption{Example images from experiment 1B where the ratings given by GPT4-V match or diverge from human psychological intuition. \textbf{A}: GPT4-V answers 7, 5, 2, 7 for left cost, right cost, left reward, right reward. These values make sense given the path of the agent. It crosses straight through the left terrain, indicating that it might have a high cost associated with it and that the agent does not find the left care package rewarding. It then crosses through the yellow terrain to pick up the white care package, indicating that this package has a high reward associated with it. \textbf{B}: GPT4-V answers 5, 4, 2, 7 for left cost, right cost, left reward, right reward. These values are counterintuitive. The agent is seen crossing into the left terrain and picking up the left reward, which indicates that the left care package has a significant reward associated with it and that the left terrain should not incurr a large cost upon the agent. GPT4-V however assigns the left terrain a higher cost than the right terrain and also the left care package a lower reward than the right care package.}
            \label{fig:psych_example_all}
        \end{figure*}

        \begin{figure*}[ht!]
            \centering
            \includegraphics[width=1.0\textwidth]{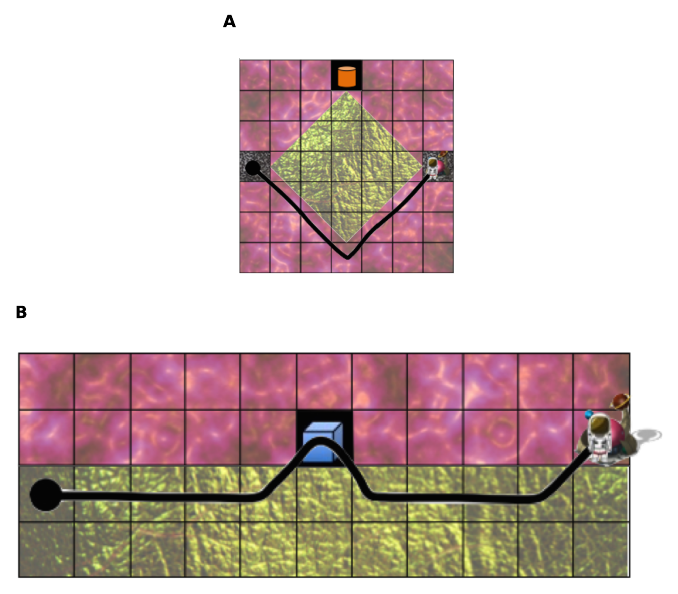}
            \caption{Example images from experiments 1A and 1B. \textbf{A}: GPT4-V answers 7, 7, 0 for inside cost, outside cost, top reward. That the care package has no value for the agent is sensible given the image, however we would expect the costs for inside and outside terrains to diverge, as the agent seemingly takes a detour to avoid crossing the inside terrain. \textbf{B}: GPT4-V answers 5, 5, 7 for bottom cost, top cost, top reward. The reward here should have a high value seeing as the agent takes a detour to collect the care package. However, the bottom terrain should be associated with a lower cost compared to the top terrain, as the agent is seen taking a detour to cross the bottom terrain instead of the top terrain after picking up the care package.}
            \label{fig:psych_example}
        \end{figure*}
    
    \clearpage

    \subsection{Intuitive psychology: Help or hinder}
    
        \begin{figure*}[ht!]
            \centering
            \includegraphics[width=1.0\textwidth]{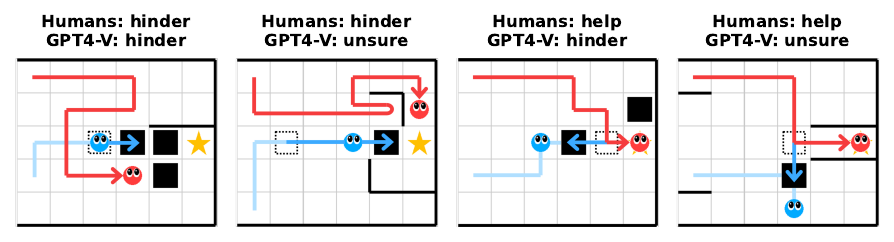}
            \caption{Example images from Wu et al. \cite{wu2023responsibility}. Humans and models were asked if the red agent would have succeeded in reaching the star if the blue agent had not been present in the scene.}
            \label{fig:psych2_example}
        \end{figure*}
    
\section{Prompting analysis}
\vspace{0.5cm}
In this additional analysis, we explore two different types of context and response constraints: the standard context and a simplified context. The standard context is close to the original task description used in the human experiment. It begins as follows: \say{This task is about astronauts. The astronauts are exploring planets with alien terrains depicted with different colours and textures [...]}. The simplified context is a modified version of the standard context that is closer to the grid level and therefore should require less abstraction from the input image: \say{The image shows a grid of squares. The white figure on the center right square is the astronaut. Each square represents a single block of terrain that the astronaut can cross in one step [...]}. 

We also explore two different types of response constraints: a short single number response and a step by step reasoning response. For the single number response, the number of output tokens is set to 3 and we begin the prompt with \say{Please answer the following question with a number only:} and end it with \say{You are only allowed to answer with a number!}. For the step by step response, we allow for 500 output tokens and begin the prompt with \say{Answer the following question and give your reasoning why:} and end it with \say{Let's think step by step and then give your final answer as a single number in the format [X].} The final single number format allows for easy machine parsing of the output. We test all four resulting combinations for GPT-4V, GPT-4V-Preview and Fuyu.

\vspace{0.5cm}
\subsection{GPT-4V}
\vspace{0.5cm}
\begin{center}
\begin{tabular}{ |c|c|c|c|c| } 
 \hline
 \textbf{Context} & \textbf{Response} & \textbf{Token length} & \textbf{Cost correlation} & \textbf{Reward correlation} \\ 
 \hline
 Standard & Single number & 3 & nan &  nan \\
 Simplified & Single number & 3 & nan & 0.4140 \\
 Standard & Step by step & 500 & -0.3008 & 0.5139 \\
 Simplified & Step by step & 500 & -0.1611 & 0.0055 \\
 \hline
\end{tabular}
\end{center}
\vspace{0.5cm}
We find that the correlation to human answers change substantially depending on the different prompting variations. For the single word binary response, the model ends up giving a constant response for most judgements (\say{7 on a scale from 1 to 10}). For the step by step reasoning, the model gives more varied answers. The highest correlation to humans is achieved for reward judgements in the standard context and the step by step response. However, this comes at the cost of a lower cost correlation with humans. 

To conclude, we find that constraining the model answers too much (for example by forcing them to output a single number) can lead to the model giving constant judgements. Instead, prompting the model to think step by step leads to more variance in the model judgements. The results on the type of context are inconclusive, as the simplified context leads to more variance in the single number response prompting, but at the same time to a lower correlation for step by step prompting.

\subsection{GPT-4V-Preview}
\vspace{0.5cm}
\begin{center}
\begin{tabular}{ |c|c|c|c|c| } 
 \hline
 \textbf{Context} & \textbf{Response} & \textbf{Token length} & \textbf{Cost correlation} & \textbf{Reward correlation} \\ 
 \hline
 Standard & Single number & 3 & nan &  0.3323 \\
 Simplified & Single number & 3 & 0.1290 & nan \\
 Standard & Step by step & 500 & nan &  0.2355 \\
 Simplified & Step by step & 500 & nan & nan \\
 \hline
\end{tabular}
\end{center}
\vspace{0.5cm}
For the previous GPT-4V model, we find that no prompting method yields good results for both cost and reward questions. 

\vspace{0.5cm}
\subsection{Fuyu}
\vspace{0.5cm}
\begin{center}
\begin{tabular}{ |c|c|c|c|c| } 
 \hline
 \textbf{Context} & \textbf{Response} & \textbf{Token length} & \textbf{Cost correlation} & \textbf{Reward correlation} \\ 
 \hline
 Standard & Single number & 3 & nan & -0.2927 \\
 Simplified & Single number & 3 & 0.1316 & -0.1858 \\
 Standard & Step by step & 500 & nan & nan \\
 Simplified & Step by step & 500 & nan & nan \\
 \hline
\end{tabular}
\end{center}
\vspace{0.5cm}
For Fuyu, we find that the correlations to human answers again change drastically depending on the different prompting variations. For the single number response, the model seems to be slightly better aligned to humans when given the simplified prompt. Crucially however, the model does not produce sensible outputs when given the instruction to perform step by step reasoning. For cost questions, it keeps repeating the following: \say{The astronaut on the center left square can cross a pink terrain easily, as it is rated 0. The astronaut on the center right square can cross a pink terrain easily, as it is rated 0. The astronaut on the center left square can cross a pink terrain, but it is rated 0. The astronaut on the center right square can cross a pink terrain, but it is rated 0 [...]}. For reward questions on the other hand, it repeats: \say{The white object on a scale of 7 (not a lot) is rewarding for the astronaut because it represents multiple blocks of terrain that can be crossed simultaneously. The white object on a scale of 6 (not a lot) is rewarding for the astronaut because it represents multiple blocks of terrain that can be crossed simultaneously. The white object on a scale of 5 (a lot) is rewarding for the astronaut because it represents multiple blocks of terrain that can be crossed simultaneously [...]}. 

\vspace{0.5cm}
\subsection{Conclusion}
\vspace{0.5cm}
We show that the prompting strategy affects behavior for both models for the intuitive psychology astronaut task. During initial testing, we tried different prompt variations and settled on a set of prompts that worked best over all models in practice. While strategies such as step by step reasoning can improve the performance of powerful models such as GPT-4V, the performance of less powerful models such as Fuyu can deteriorate (likely due to the increased prompt complexity). 
    
Additionally, we found that models can be impacted by small changes to prompts on a character level, likely due to the effects of tokenization. For example, in the counterfactual task for the Michotte experiment, we query Fuyu with \say{Q: The scene shows two balls labeled “A”, and “B”. On the left side there is a pink gate. The solid arrows show the trajectories of the balls. Indicate your agreement with the assessment that ball “B” would have gone through the gate if ball “A” had not been present in the scene on a scale from 0 (not at all) to 100 (very much). You are only allowed to answer with a single number.}. If we instead introduce the slight variation of removing the quotation marks around A and B in the second to last sentence: \say{[...] Indicate your agreement with the assessment that ball B would have gone through the gate if ball A had not been present [...]}, the ratings given by Fuyu switch completely for some sequences (from 0 to 100 or 100 to 0). While this small change heavily impacts Fuyu's behavior, GPT-4V-Preview only switches from 100 to 0 for one sequence, and Claude-3 only varies ratings between 90 and 100. Taken together these results not only indicate that model behavior can be sensitive to small changes to prompts on a character level but also that the behavior of more capable models might be less impacted. 

\end{appendices}
\end{document}


\nobibliography{bibliography}
\thispagestyle{empty}
\vspace{-3cm}
\section*{Supplementary information}
\vspace{1cm}

\section*{Prompts}\label{secA1}

    \subsection*{Intuitive physics: Block towers}
    For the intuitive physics experiment, we used a task from Lerer et al. \cite{lerer2016learning}. For each trial, we asked the models three questions: 
        
        \begingroup
            \setlength{\tabcolsep}{6pt} 
            \renewcommand{\arraystretch}{3}
            \begin{table}[!h]
            \begin{tabularx}{\linewidth}{ c X } 
            \hline
            Question 1 & What is the background color? \\ \hline
            Question 2 &  What are the colors of the blocks from top to bottom? You are only allowed to answer with the color names. \\ \hline
            Question 3 & Will this block tower fall? Give a boolean answer. \\ \hline
            \end{tabularx}
            \end{table}
        \endgroup

    \subsection*{Causal reasoning: Jenga}
    For the first causal reasoning experiment, we used a task from Zhou et al. \cite{zhou2023mental, gerstenberg2017faulty}. For each trial, we again asked the models three questions: 
        
        \begingroup
            \setlength{\tabcolsep}{6pt} 
            \renewcommand{\arraystretch}{3}
            \begin{table}[!h]
            \begin{tabularx}{\linewidth}{ c X } 
            \hline
            Question 1 & How many blocks are there in the image? You are only allowed to answer with a single number. No words allowed! \\ \hline
            Question 2 & How many of the red bricks would fall off the table if the dark grey brick wasn't there? You are only allowed to answer with a single number between 0 and \{num\_blocks\} corresponding to how many blocks would fall. No words allowed! \\ \hline
            Question 3 & How responsible is the dark grey brick for the red bricks staying on the table? You are only allowed to answer with a number on a scale from 0\% (not at all responsible) to 100\% (fully responsible). No words allowed! \\ \hline
            \end{tabularx}
            \end{table}
        \endgroup

    \newpage
    \subsection*{Causal reasoning: Michotte}
    For the second causal reasoning experiment, we used a task from Gerstenberg et al. \cite{gerstenberg2017eye}. For each trial, we asked the models four questions: 
        
        \begingroup
            \setlength{\tabcolsep}{6pt} 
            \renewcommand{\arraystretch}{3}
            \begin{table}[!h]
            \begin{tabularx}{\linewidth}{ c X } 
            \hline
            Question 1 &  The scene shows two balls labeled “A”, and “B”. On the left side there is a pink gate. The solid arrows show the trajectories of the balls. What is the background color of the image? You are only allowed to answer with a single color name! \\ \hline
            Question 2 & The scene shows two balls labeled “A”, and “B”. On the left side there is a pink gate. The solid arrows show the trajectories of the balls. Do the balls move “left to right” or “right to left”? You are only allowed to respond with one of the two options! \\ \hline
            Question 3 & The scene shows two balls labeled “A”, and “B”. On the left side there is a pink gate. The solid arrows show the trajectories of the balls. Indicate your agreement with the assessment that ball “B” [completely missed / went through the middle of] the gate on a scale from 0 (not at all) to 100 (very much). You are only allowed to answer with a single number. \\ \hline
            Question 4 & The scene shows two balls labeled “A”, and “B”. On the left side there is a pink gate. The solid arrows show the trajectories of the balls. Indicate your agreement with the assessment that ball “B” would have gone through the gate if ball “A” had not been present in the scene on a scale from 0 (not at all) to 100 (very much). You are only allowed to answer with a single number. \\ \hline
            \end{tabularx}
            \end{table}
        \endgroup
        
    \subsection*{Intuitive psychology: Astronaut}
    For the first intuitive psychology experiment, we used three tasks from Jara-Ettinger et al. \cite{jara2020naive}. We again asked the model two descriptive questions in order to assess their basic comprehension of the scene:

        \begingroup
            \setlength{\tabcolsep}{6pt} 
            \renewcommand{\arraystretch}{3}
            \begin{table}[!h]
            \begin{tabularx}{\linewidth}{ c X } 
            \hline
            Question 1 & Please answer the following question with a single color name only: What color is the background of the central part of the image? You are only allowed to answer with a single color name!\\ \hline
            Question 2 & Please answer the following question with a number only: How many orange or white containers are in the image? You are only allowed to answer with a number! \\ \hline
            \end{tabularx}
            \end{table}
        \endgroup
    
    \clearpage
    \noindent For experiment 1A we first gave the models the following basic prompt, which was combined with different trial specific questions: 
    
    \begin{quote}
        \say{This task is about astronauts. The astronauts are exploring planets with alien terrains depicted with different colours and textures. Each astronaut has different skills, making each terrain more or less exhausting or easy for them to cross. All astronauts can ultimately cross all terrains, even if it's exhausting. The astronauts land far from the base and have to walk there. In each image, the black circle on the left indicates where the astronaut landed. The base is on the middle right part of the image. Sometimes care packages are dropped from above and the astronauts can pick them up. There are two kinds of care packages depicted with an orange cylinder and a white cube. Each astronaut has different preferences and likes each kind of care package in different amounts. The astronauts don't actually need the care packages. They can go straight to the base, or they can pick one up. You will see images of different astronauts with different skills and preferences travelling from their landing location to the home base. The astronauts always have a map. So they know all about the terrains and the care packages. Please answer the following question with a number only:}
        \\
    \end{quote}

    \noindent For experiment 1B we first gave the models the following basic prompt, which was again combined with different trial-specific questions: 
    
    \begin{quote}
        \say{This task is about astronauts. The astronauts are exploring planets with alien terrains depicted with different colours and textures. Each astronaut has different skills, making each terrain more or less exhausting or easy for them to cross. All astronauts can ultimately cross all terrains, even if it's exhausting. Sometimes, the astronauts land far from the base and have to walk there. In each image, the black circle indicates where the astronaut landed. The base is in the center of the image. Sometimes care packages are dropped from above and the astronauts can pick them up. There are two kinds of care packages depicted with an orange cylinder and a white cube. Sometimes both care packages are identical. The astronauts cannot pick both care packages. Each astronaut has different preferences and likes each kind of care package in different amounts. The astronauts don't actually need the care packages. They can go straight to the base, or they can pick one up. You will see images of different astronauts with different skills and preferences travelling from their landing location to the home base. The astronauts always have a map. So they know all about the terrains and the care packages. Please answer the following question with a number only:}
        \\
    \end{quote}

    \noindent Finally, for experiment 1C we first gave the models the following basic prompt: 
    
    \begin{quote}
        \say{This task is about astronauts. The astronauts are exploring planets with alien terrains depicted with different colours and textures. Each astronaut has different skills, making each terrain more or less exhausting or easy for them to cross. All astronauts can ultimately cross all terrains, even if it's exhausting. The astronauts land far from the base and have to walk there. In each image, the black circle on the left indicates where the astronaut landed. The base is on the right part of the image. The path astronauts take from where they land to their base is indicated by a thick black line between the black circle on the left and the astronaut on the right. Sometimes care packages depicted by a blue cube on a black background are dropped from above and the astronauts can pick them up. Each astronaut has different preferences and likes each care package in different amounts. The astronauts don't actually need the care packages. They can go straight to the base, or they can pick one up. You will see images of different astronauts with different skills and preferences travelling from their landing location to the home base. Your task is to judge how easy/exhausting it is for the astronaut in each image to cross each terrain, and how much they like each care package. The astronauts always have a map. So they know all about the terrains and the care packages. Please answer the following question with a number only:}
        \\
    \end{quote}

    \clearpage
    \subsubsection*{Experiment 1A}

    \begingroup
        \setlength{\tabcolsep}{6pt} 
        \renewcommand{\arraystretch}{2}
        \renewcommand\tabularxcolumn[1]{m{#1}}
        \begin{table}[!h]
        \begin{tabularx}{\linewidth}{ c X } 
        \textbf{Images} & \textbf{Questions} \\ \hline
        0, 1 & 
        \vspace{0.4cm}
        \begin{enumerate}
        \item How easy is it for the astronaut in this image to cross the pink terrain on a scale from 0 (extremely easy) to 10 (extremely exhausting)? You are only allowed to answer with a number! 
        \item How much does the astronaut in this image like the orange care package on a scale from 0 (not at all) to 10 (a lot)? You are only allowed to answer with a number! 
        \end{enumerate}\\ \hline
        2, 3, 4, 5 &
        \vspace{0.4cm}
        \begin{enumerate}
        \item How easy is it for the astronaut to cross the pink terrain on a scale from 0 (extremely easy) to 10 (extremely exhausting)? You are only allowed to answer with a number! \item How much does the astronaut like the white care package on a scale from 0 (not at all) to 10 (a lot)? You are only allowed to answer with a number! 
        \item How much does the astronaut like the orange care package on a scale from 0 (not at all) to 10 (a lot)? You are only allowed to answer with a number!"
        \end{enumerate}\\ \hline
        6, 7, 8, 9, 10 &
        \vspace{0.4cm}
        \begin{enumerate}
        \item How easy is it for the astronaut to cross the purple terrain on a scale from 0 (extremely easy) to 10 (extremely exhausting)? You are only allowed to answer with a number! \item How easy is it for the astronaut to cross the pink terrain on a scale from 0 (extremely easy) to 10 (extremely exhausting)? You are only allowed to answer with a number! \item How much does the astronaut like the orange care package on a scale from 0 (not at all) to 10 (a lot)? You are only allowed to answer with a number! 
        \end{enumerate} \\ \hline
        11, 12, 13, 14, 15 &
        \vspace{0.4cm}
        \begin{enumerate}
        \item How easy is it for the astronaut to cross the purple terrain on a scale from 0 (extremely easy) to 10 (extremely exhausting)? You are only allowed to answer with a number! \item How easy is it for the astronaut to cross the pink terrain on a scale from 0 (extremely easy) to 10 (extremely exhausting)? You are only allowed to answer with a number! \item How much does the astronaut like the white care package on a scale from 0 (not at all) to 10 (a lot)? You are only allowed to answer with a number!
        \item How much does the astronaut like the orange care package on a scale from 0 (not at all) to 10 (a lot)? You are only allowed to answer with a number!" 
        \end{enumerate} \\ \hline
        \end{tabularx}
        \end{table}
    \endgroup

    \clearpage
    \subsubsection*{Experiment 1B}

    \begingroup
        \setlength{\tabcolsep}{6pt} 
        \renewcommand{\arraystretch}{2}
        \renewcommand\tabularxcolumn[1]{m{#1}}
        \begin{table}[!h]
        \begin{tabularx}{\linewidth}{ c X } 
        \textbf{Images} & \textbf{Questions} \\ \hline
        1, 2, 3, 4, 5, 6, 7 & 
        \vspace{0.4cm}
        \begin{enumerate}
        \item How easy is it for the astronaut to cross the pink terrain on a scale from 0 (extremely easy) to 10 (extremely exhausting)? You are only allowed to answer with a number! \item How easy is it for the astronaut to cross the yellow terrain on a scale from 0 (extremely easy) to 10 (extremely exhausting)? You are only allowed to answer with a number! \item How much does the astronaut like the orange care package on a scale from 0 (not at all) to 10 (a lot)? You are only allowed to answer with a number! \item How much does the astronaut like the white care package on a scale from 0 (not at all) to 10 (a lot)? You are only allowed to answer with a number!
        \end{enumerate}\\ \hline
        8, 9, 10 &
        \vspace{0.4cm}
        \begin{enumerate}
        \item How easy is it for the astronaut to cross the pink terrain on a scale from 0 (extremely easy) to 10 (extremely exhausting)? You are only allowed to answer with a number! \item How easy is it for the astronaut to cross the yellow terrain on a scale from 0 (extremely easy) to 10 (extremely exhausting)? You are only allowed to answer with a number! \item How much does the astronaut like the orange care package on a scale from 0 (not at all) to 10 (a lot)? You are only allowed to answer with a number!
        \end{enumerate}\\ \hline
        11, 12, 13, 14, 15, 16, 17 &
        \vspace{0.4cm}
        \begin{enumerate}
        \item How easy is it for the astronaut to cross the pink terrain on a scale from 0 (extremely easy) to 10 (extremely exhausting)? You are only allowed to answer with a number! \item How much does the astronaut like the orange care package on a scale from 0 (not at all) to 10 (a lot)? You are only allowed to answer with a number! \item How much does the astronaut like the white care package on a scale from 0 (not at all) to 10 (a lot)? You are only allowed to answer with a number!
        \end{enumerate} \\ \hline
        \end{tabularx}
        \end{table}
    \endgroup

    \subsubsection*{Experiment 1C}

    \begingroup
        \setlength{\tabcolsep}{6pt} 
        \renewcommand{\arraystretch}{2}
        \renewcommand\tabularxcolumn[1]{m{#1}}
        \begin{table}[!h]
        \begin{tabularx}{\linewidth}{ c X } 
        \textbf{Images} & \textbf{Questions} \\ \hline
        All images & 
        \vspace{0.4cm}
        \begin{enumerate}
        \item How easy is it for the astronaut to cross the yellow terrain on a scale from 0 (extremely easy) to 10 (extremely exhausting)? You are only allowed to answer with a number! \item How easy is it for the astronaut to cross the pink terrain on a scale from 0 (extremely easy) to 10 (extremely exhausting)? You are only allowed to answer with a number! \item How much does the astronaut like the blue care package on a scale from 0 (not at all) to 10 (a lot)? You are only allowed to answer with a number!
        \end{enumerate}\\ \hline
        \end{tabularx}
        \end{table}
    \endgroup

    \subsection*{Intuitive psychology: Agents}
    For the second intuitive psychology experiment, we used a task from Wu et al. \cite{wu2023responsibility}. For each trial, we gave the models the following basic prompt, followed by one of the four questions: 

      \begin{quote}
         \say{The scene shows two agents in a grid world in which agents and objects can interact. On each timestep, agents can move up, down, left, right, or stay in place, but cannot move through walls or boxes. One agent, RED, has a physical goal of reaching a star in 10 timesteps. If they run out of time, then they fail. Another agent, BLUE, has a social goal of helping or hindering RED. BLUE has the ability to push or pull boxes around.}
        \\
    \end{quote}
        
        \begingroup
            \setlength{\tabcolsep}{6pt} 
            \renewcommand{\arraystretch}{3}
            \begin{table}[!h]
            \begin{tabularx}{\linewidth}{ c X } 
            \hline
            Question 1 &  What is the background color of the image? You are only allowed to answer with a single color name! \\ \hline
            Question 2 & How many boxes are in the scene? You are only allowed to respond with a single number. \\ \hline
            Question 3 & What was BLUE intending to do? Give your answer on a scale from “definitely hinder RED” (0) to “definitely help RED” (100) with the midpoint “unsure” (50). You are only allowed to respond with a single number. \\ \hline
            Question 4 & How much do you agree that RED would have (still) succeeded if BLUE hadn't been there on a scale from “not at all” (0) to “very much” (100)? You are only allowed to answer with a single number. \\ \hline
            \end{tabularx}
            \end{table}
        \endgroup